\documentclass[letterpaper, 10 pt, journal, twoside]{IEEEtran}
\usepackage{amsmath, amssymb, amsfonts}
\usepackage{algorithmicx, algorithm}
\usepackage{amsthm}
\usepackage{array}
\usepackage{textcomp}
\usepackage{stfloats}
\usepackage{verbatim}
\usepackage{graphicx}
\usepackage{subfigure}
\usepackage{picinpar}
\usepackage{url}
\usepackage{flushend}
\usepackage{colortbl}
\usepackage{soul}
\usepackage{multirow}
\usepackage{pifont}
\usepackage{alltt}
\usepackage{float}
\usepackage{enumerate}
\usepackage{siunitx}
\usepackage{breakurl}
\usepackage{epstopdf}
\usepackage{pbox}
\usepackage{booktabs}
\usepackage{makecell}
\usepackage{threeparttable}
\usepackage{balance}
\usepackage{caption}
\usepackage{gensymb}
\usepackage{bm}
\usepackage{fancyhdr}
\usepackage[table]{xcolor}
\usepackage{times} % Times字体，不包含数学字体
\usepackage{relsize}

\definecolor{mygray}{gray}{.9}
\definecolor{mypink}{rgb}{.99,.85,.85}
\definecolor{mycyan}{cmyk}{.3,0,0,0}

\usepackage{algpseudocode}
\usepackage{hyperref}
\hypersetup{
    colorlinks=true,
    linkcolor=black,
    citecolor=black,
    urlcolor=blue,
    pdfborder={0 0 0}
}

\begin{document}   %正式开始这个文档
\title{MILD: Tractable Terrain Modeling for Learning Improved Bipedal Locomotion on Deformable Surfaces}

% \author{Weipeng Guan$^{1}$ and Peng Lu$^{2}$% <-this % stops a space
\author{Zeren Luo$^{1}$,  Jiahui Zhang$^{1}$,  Zhe Xu$^{2}$,  Wanyue Li$^{1}$, Xinqi Li$^{1}$, Xuechao Chen$^{2}$, Zhangguo Yu$^{2}$, Annan Tang$^{3, \dag}$, Peng Lu$^{1, \dag}$
\vspace{-1.7em}

% \thanks{*This work was supported by *******}% <-this % stops a space
% \thanks{$^{1}$Guan Weipeng is with Faculty of Engineering, Department of Mechanical Engineering, The University of Hong Kong, SAR Hong Kong, China
%         {\tt\small guanwp@hku.hk}}%
% \thanks{$^{2}$Lu Peng is with Faculty of Engineering, Department of Mechanical Engineering, The University of Hong Kong, SAR Hong Kong, China
%         {\tt\small lupeng@hku.hk}}%
\thanks{
    Manuscript received: April 28, 2025; Revised: August 30, 2025; Accepted: December 4, 2025.
    
    This paper was recommended for publication by Editor Aleksandra Faust upon evaluation of the Associate Editor and Reviewers’ comments. This work was supported by General Research Fund under Grant No. 17204222. ($^{\dag}$Corresponding author \url{lupeng@hku.hk}, \url{tang@jsk.imi.i.u-tokyo.ac.jp}).

    $^{1}$The authors are with the Adaptive Robotic Controls Lab (ArcLab), Department of Mechanical Engineering, The University of Hong Kong, Hong Kong.

    $^{2}$The authors are with School of Mechanical and Electrical Engineering, Beijing Institute of Technology, Beijing, China.

    $^{3}$The authors is with Graduate School of Information Science and Technology, The University of Tokyo, Tokyo, Japan.

    Digital Object Identifier (DOI): see top of this page.
}%
}

\markboth{IEEE ROBOTICS AND AUTOMATION LETTERS.~PREPRINT VERSION.~ACCEPTED VERSION}%
{LUO\&ZHANG \MakeLowercase{\textit{et al.}}: MILD: Tractable Terrain Modeling for Learning Improved Bipedal Locomotion on Deformable Surfaces}

% \IEEEpubid{0000--0000/00\$00.00~\copyright~2015 IEEE}

% \IEEEpubid{\begin{minipage}{\textwidth}\ \\[30pt] \centering
% 		Copyright \copyright 2023 IEEE. Personal use of this material is permitted. 
% 		However, permission to use this material for any other purposes must \\ be obtained 
% 		from the IEEE by sending an email to pubs-permissions@ieee.org.
% \end{minipage}}

\maketitle  %不要加页面，避免编译不过
% \thispagestyle{headings} 
% \pagestyle{headings} %添加这个让每页都有   % empty - 没有页眉和页脚  % plain - 没有页眉，页脚包含一个居中的页码  % headings - 没有页脚，页眉包含章/节或者字节的名字和页码 % myheadings - 没有页脚，页眉包含有页码

%%%%%%%%%%%%%%%%%%%%%%%%%%%%%%%%%%%%%%%%%%%%%%%%%%%%%%%%%%%%%%%%%%%%%%%%%%%%%%%%
\begin{abstract}
Enabling robots to walk on yielding terrain is vital for applications ranging from disaster response to planetary exploration. While bipedal robots hold immense potential, their locomotion on deformable surfaces remains limited as current simulators fail to capture the spatiotemporal heterogeneity of such yielding substrates. 
We present MILD, featuring a physics-grounded discrete-element contact solver that accurately simulates spatially varying foot-terrain interactions. Complementing this model, we train a terrain-aware locomotion controller via deep reinforcement learning with latent modulation and proprioceptive estimation.
Quantitative comparisons against state-of-the-art methods show our approach generates more diverse and realistic contact scenarios during training, resulting in controllers that exhibit natural adaptation on real deformable surfaces. Through hardware experiments, we demonstrate the system's capability for online terrain identification and adaptation across a wide range of surface stiffness.

\end{abstract}

%%%%%%%%%%%%%%%%%%%%%%%%%%%%%%%%%%%%%%%%%%%%%%%%%%%%%%%%%%%%%%%%%%%%%%%%%%%%%%%%

%%%%%%%%%%%%%%%%%%%%%%%%%%%%%%%%%%%%%%%%%%%%%%%%%%%%%%%%%%%%%%%%%%%%%%%%%%%%%%%%
% ***************************************************************************
% ***************************************************************************
% ****************************** S E C T I O N ******************************
% ***************************************************************************
% ***************************************************************************
\section{INTRODUCTION}
\label{INTRODUCTION}

\IEEEPARstart
{I}{n} recent years, a notable upsurge has been witnessed in the bipedal robotics sector, as it shows great potential for application in deformable and unstructured terrain, which constitutes a significant portion of the Earth's surface. 
Current bipedal robot locomotion controllers are predominantly optimized for rigid ground,
as mainstream simulators \cite{todorov2012mujoco} \cite{coumans2021} \cite{makoviychuk2021isaac} rely on rigid-body dynamics that cannot capture the spatiotemporal heterogeneity of deformable terrains. This fundamental limitation prevents accurate modeling of soft-surface interactions, ultimately restricting the generation of realistic data for controller design.

\begin{figure}
   \centering
   \vspace{-1.0em}
    \includegraphics[width=0.49\textwidth, trim=1 1 1 1,clip]{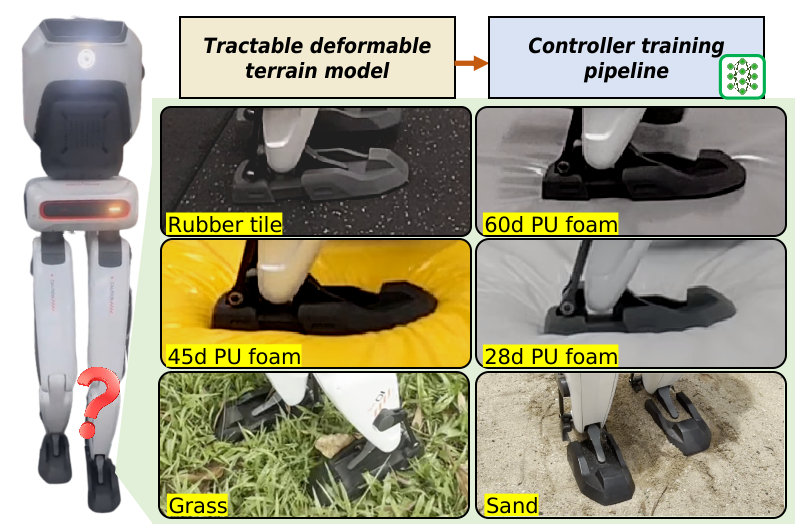}
   \caption{ The bipedal robot as a multi-articulated floating-base system interacting with compliant surfaces through integrated physics-grounded modeling and learning-based control. The locomotion is demonstrated across terrains with parametrically varied stiffness and deformability (d -- density of the polyurethane foam. Unit: $kg/m^3$).}
   \label{pics:cover}
   \vspace{-1.5em}
\end{figure}

\subsection{Soft Surface Simulation and Modeling} 
In the robotics physics community, exploring methods that can accurately and efficiently simulate the deformation and stress of soft substrates upon contact with robots has always been a highly valuable research direction.
The study begins with granular media (GM), complex systems composed of discrete particles. 
Particle-based simulations can model grain interactions but are computationally expensive, often requiring days to simulate just one second of real-time behavior \cite{poschel2005computational}.
To elucidate the mechanical principles of GM interactions with robotic entities, researchers propose simplified contact mechanics such as linear viscoelastic models \cite{lynch2020soft, lynch2024efficient} or rate-dependent plastic models \cite{ding2013foot, vasilopoulos2014compliant}. While computationally efficient, these models oversimplify the hydrodynamic-like nature of GM, leading to significant deviations from real-world behavior.
For a more accurate description of GM mechanics, the resistive force theory (RFT) is introduced \cite{li2009sensitive, li2013terradynamics}, which is an empirical model that describes quasistatic reaction forces. It is applicable to biped quasistatic locomotion when combined with a ZMP-based controller \cite{xiong2017stability}. The theory maintains its validity when extended to three-dimensional robots intrusion scenarios with GM \cite{treers2021granular, chen2024foot}.

Despite this, RFT is primarily applicable to slow interactions failing to capture reaction forces during high-dynamic penetration. For high-speed penetration, researchers propose impulsive force models \cite{katsuragi2007unified, tsimring2005modeling}, simplifying forces as instantaneous actions. Another study reveals limitations in these models, as high-speed penetration compresses and accumulates grains beneath the object, inducing an added mass effect that generates additional forces on the contact surface \cite{aguilar2016robophysical}. This developing cone principally reduces the formulation, and its tractable solution enhances computational efficiency in simulations. 
In this work, we extend it to a more generalized model, taking into account scenarios involving multiple target areas' interactions, which are overlooked in the original model.

% - 这个 granular cone model只研究了虽然computationally efficient,但面对general locomotion task不够，我们采用了 \cite{}的方法extend了tangential force。
\subsection{Legged Locomotion on Deformable Terrain} 
% While model-based controllers have successfully implemented whole-body dynamics optimization for rigid terrain locomotion \cite{di2018dynamic, meng2023online}, these methods break down when applied to deformable substrates.
% The fundamental challenge lies in the coupled dynamics between the robot and continuously deforming terrain, which invalidates the static contact assumptions underlying current approaches.
% Therefore, most advancements in this field have been restricted to reduced-degree-of-freedom systems, such as unidirectional hoppers \cite{hubicki2016tractable, lynch2020soft, aguilar2016robophysical}, bipeds with support beams \cite{xiong2017stability}, or wheeled platforms with fixed upper bodies \cite{huang2022dynamic} - leaving more complex legged systems largely unexplored.
\textcolor{black}{Early research in this field is largely restricted to reduced-degree-of-freedom systems, such as unidirectional hoppers \cite{hubicki2016tractable, lynch2020soft, aguilar2016robophysical} or simplified bipedal models \cite{xiong2017stability}, where the limited kinematic complexity enables tractable analysis and control design.
Conventional model-based controllers fundamentally struggle with the coupled dynamics between the robot and deforming surface - particularly the violation of static contact assumptions that underpin their stability guarantees \cite{di2018dynamic}. 
Passivity-based control frameworks \cite{henze2016passivity, mesesan2019dynamic} resolve this limitation through an impedance wrench that regulates the center of mass and end-effector trajectory, enabling stable whole-body locomotion on compliant terrain.}

Recent advances in reinforcement learning enable legged robots to traverse various challenging terrains, \textcolor{black}{including deformable surfaces like grass and sand \cite{lee2020learning, radosavovic2024learning}. These methods often combine implicit system identification with domain randomization over rigid terrain geometries to achieve generalization.  \cite{singh2024robust} specifically simulates terrain compliance by randomizing parameters of MuJoCo's built-in mass-spring-damper contact model.  
While certain parameter combinations may partially capture characteristics of soft substrate interactions, this approximation remains fundamentally limited compared to real-world terrain penetration. }
% During deployments, significant foot sinking into soft ground deviates substantially from training conditions. 
The learned controllers's performance deteriorates considerably due to the lack of exposure to such data. 

Recent studies attempt to establish more realistic models of yielding terrain in simulation for quadrupeds, aiming to provide higher-fidelity training data for RL-based controllers and bridge the reality gap \cite{chen2024identifying, choi2023learning}. A key limitation in these works is their assumption that the reaction force is uniformly distributed across the contact area, which disregards the influence of insertion posture on contact force computation \cite{ding2013foot}. However, for bipedal robots, this assumption becomes invalid due to their large footplates, where contact forces vary significantly across different regions. 
In this work, the aim is to solve the aforementioned issues regarding the modeling accuracy and efficiency in an RL formulation. We propose \textbf{MILD}: a tractable \textbf{M}odel achieving \textbf{I}mproved bipedal \textbf{L}ocomotion on \textbf{D}eformables.

In summary, our walking controller introduces innovations in threefold: (i) We propose a tractable foot-terrain interaction model for substantial contact areas, overcoming the prevailing assumption of uniform force distribution in prior work (e.g., quadrupeds and 1-D hoppers). This model explicitly accounts for eccentric insertion and spatiotemporal heterogeneous penetration depths, enabling accurate force prediction for biped-scale footplates. (ii) Building on this, we develop a novel RL-based training framework with an encoder-modulation architecture that implicitly adapts to terrain compliance by dynamically encoding the properties into latent representations. (iii) Through extensive simulations and hardware validations, we demonstrate that the proposed model outperforms state-of-the-art methods in robustness and energy efficiency, and the controller's capability of online identifying and handling abrupt terrain stiffness transitions.

% ***************************************************************************
% ***************************************************************************
% ****************************** S E C T I O N ******************************
% ***************************************************************************
% ***************************************************************************
\section{METHOD}\label{methodology}
% \vspace{-0.1em}
Generalized contact solver for robot-deformable terrain interactions extends physics-grounded models through discrete-element segmentation with continuity constraints. Integrated into the simulators, it enables RL-based bipedal locomotion training. 
Section \ref{GM_model} - \ref{GM_solver} illustrates the contact solver’s derivation, and Section \ref{sec:training} details the RL problem formulation.
%我们的工作主要集中在推导一种机器人与deformable surface的接触模型和接触求解器，基于一个被实验证明的物理学principle。但是我们对其做了理论推广但同时满足一系列新的连续性约束，使其能适应到更为一般的接触情况。这种模型被integrate到一个商业simluation中训练RL-based的bipedal robot 的locomotion。我们将在following section介绍contact solver的推导过程以及RL problem的formulation。

\begin{figure*}[h]
   \centering
    \includegraphics[width=1.005\textwidth, trim=8 0 0 0,clip]{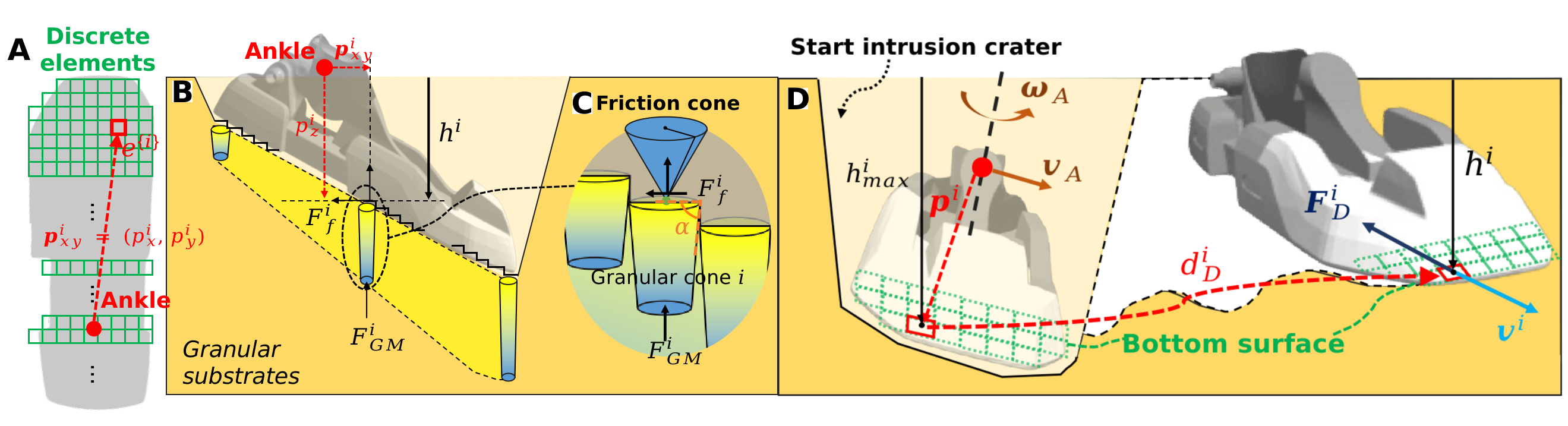}
   \caption{ Eccentric penetration mechanics of granular media (GM) under arbitrary-shaped feet. The footplate is discretized into elements (DEs), each modeling localized GM interactions via independent developing cones (Eq. (\ref{eq:flat}) - (\ref{eq:fgm})). 
   Deeper intrusion amplifies the cone’s added mass effect. The model requires each element to satisfy individual friction cone constraints while maintaining continuity constraints between element pairs (Eq. (\ref{eq:delta_rb}) - (\ref{eq:fric_combined})).
   Drift-resistive forces are opposite to the element-wise motion and are scaled by the accumulated post-intrusion moving distance.    }
   \label{pics:terr_model}
   \vspace{-1.0em}
\end{figure*}

% In addition, we also define a temporal observation $\boldsymbol{o}_t^H= \left[ \boldsymbol{o}_t, \boldsymbol{o}_{t-1}, ..., \boldsymbol{o}_{t-H} \right]$ and a privileged morphology observation $\boldsymbol{o}^{mor}$, where $\boldsymbol{o}_t^H$ denotes the history of state and actions over the past $H$ time steps ($H=5$ in this task), and $\boldsymbol{o}^{mor}$ represents the masses and sizes of the agent's trunk and legs.

% ***********************************************************************
% ****************************** SECTION 2 ******************************
% ***********************************************************************
\subsection{Discrete Element-based Granular Media Model} \label{GM_model} 
% In \cite{aguilar2016robophysical}, the varying terrain stiffness arises from depth-dependent flat-surface forces and conical-surface forces acting on a developing cone, but only for circular contact surfaces. 
% Subsequent work \cite{choi2023learning} shows that even with rectangular contact approximations, the jamming cone model still reliably predicts terrain deformable behavior.
% Nevertheless, these studies share a key limitation: 
% They assume axisymmetric penetration and uniformly distributed reaction forces due to examining only small robot foot areas.
\textcolor{black}{Prior work models terrain stiffness via depth-dependent forces on a developing cone, initially for circular contacts \cite{aguilar2016robophysical} and later extends to rectangular approximations \cite{choi2023learning}. However, both assume axisymmetric penetration and uniform force distribution—limitations that stem from analyzing only small foot areas.}

\begin{figure}[H]
   \centering
    \includegraphics[width=0.5\textwidth, trim=1 1 1 1,clip]{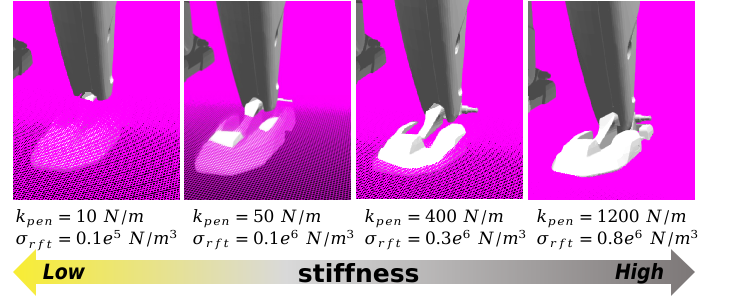}
   \caption{ Illustration of different penetration stiffness and depth-dependent resistive stress that lead to different foot penetration scenarios. The color grids indicate the terrain surface $h_0$. }
   \label{pics:terr_prop}
   \vspace{-1.5em}
\end{figure}

% For the scenario of bipedal locomotion, accurately simulating contact force distribution requires discretizing the feet envelope surface (Fig. \ref{pics:terr_model}(A)), with each element's size consistent with the scale assumptions of the aforementioned interaction models.
% This configuration ensures practical scalability for large-scale reinforcement learning training without sacrificing the diversity of terrain interactions.
% Each discrete element is treated as experiencing independent penetration, where variations in penetration depth produce distinct jamming cone geometries. This naturally produces non-uniform force distributions across the contact area.
\textcolor{black}{To accurately capture contact force distribution in bipedal locomotion, we discretize the foot envelope surface into elements (Fig. \ref{pics:terr_model}(A)) whose size aligns with granular interaction assumptions.}
\textcolor{black}{Based on preliminary experiments, a resolution of 154 elements (0.0195 m per unit) optimally balances contact richness and computational efficiency} 
\footnote{\textcolor{black}{Beyond this resolution, the marginal gain in contact richness is outweighed by a prohibitive increase in computation time, which severely compromises training scalability.}
}
.
\textcolor{black}{Each element undergoes independent penetration, forming distinct jamming cones that naturally yield non-uniform force distributions.}
For each penetrating element $e^{\{i\}}$, a growing compaction cone forms in the granular medium. The contact area consists of two components:
\begin{equation} \label{eq:flat}
    S^{\{i\}}_{flat}(h^{\{i\}}) = \pi (R - \nu \frac{h^{\{i\}} - h^{\{i\}}_{0}}{\tan \alpha})^{2}
\end{equation}
\begin{equation} \label{eq:cone}
    S^{\{i\}}_{cone}(h^{\{i\}}) = \frac{\pi R^{2} - S_{flat}(h^{\{i\}})}{\cos \alpha},
\end{equation}
\textcolor{black}{where $h^{\{i\}}$ is the penetration depth, $\alpha$ is the shear band angle, and $\nu$ is the recruitment rate. $R$ is the span of the discrete element at the intersected point $h^{\{i\}}_{0}$ (subscript $\{i\}$ is omitted in the following equations for brevity).} Added-mass $m_{A}$ describes the accumulated effect of the compacted grains inside the jammed cone, which is given as follows:
\begin{equation} \label{ma}
    m_{A} = -c_{g} \phi  \rho \nu \int_{h_{0}}^{h} S_{flat}(z) dz,
\end{equation}
where $\phi$ is the volume fraction, which indicates the degree of looseness in the interparticle connections. $\rho$ and $c_{g}$ are the grain density and surrounding mass scaling factor, respectively.
Granular reaction force $F_{GM}$ in the normal direction is composed of quasistatic force $F_g$ and the force induced by the momentum change caused by collision with the virtual added mass $d(m_{A}\dot{h})$.
% \begin{align}
% F_{GM} &= F_p(z) - c_d \frac{d m_A}{dt} \dot{h} - m_a \ddot{h} \\
% &= \frac{k_{sh}}{s^i} \int_0^{h_i} A_{flat}(h) dh + \sigma_{cone} \int_0^{h_i} A_{cone}(h) dh \nonumber \\
% &\hfill - c_d \dot{m}_a \dot{h} - m_A \ddot{h}
% \end{align}
\begin{equation} \label{eq:fg}
    F_g = \frac{k_{pen}}{s^{\{i\}}} \int_{h_{0}}^{h} S_{flat}(z) dz + \sigma_{cone} \int_{h_{0}}^{h} S_{cone}(z) dz
\end{equation}
\vspace{-1.5em} % 向上回退一行的距离
\begin{align} \label{eq:fgm}
F_{GM} &= F_g - c_d \frac{d m_A}{dt} \dot{\boldsymbol{v}}_z - m_{A} \ddot{\boldsymbol{v}}_z 
\end{align}
where $k_{pen}$ is the penetration stiffness, $c_d$ is the inertial drag scaling factor and \textcolor{black}{$\sigma_{cone}$ is the depth-dependent conical stress.} Differential added mass $\frac{d m_A}{dt} \dot{\boldsymbol{v}}_z$ can be represented as $\frac{d m_A}{dh} \dot{\boldsymbol{v}}_z^{2} = -c_{g} \phi  \rho \nu S_{flat} \dot{\boldsymbol{v}}_z^{2}$ according to the chain rule.
Besides, the velocity of $e^{\{i\}}$ w.r.t the world frame is accessible given the ankle's velocity $\boldsymbol{v}_{A}$, $\boldsymbol{\omega}_{A}$, and the vector from the ankle point to $e^{\{i\}}$, i.e. $\boldsymbol{v}^{\{i\}} = \boldsymbol{v}_{A} +\boldsymbol{\omega}_{A} \times \boldsymbol{p}^{\{i\}}$. 
% \begin{align}
%     \boldsymbol{v}^{\{i\}} = \boldsymbol{v}_{A} +\boldsymbol{\omega}_{A} \times \boldsymbol{p}^{\{i\}}
% \end{align}
It is worthwhile to note that the two integrations of Eqs. (\ref{eq:fg} - \ref{eq:fgm}) possess closed-form expressions as functions of $h$ and $h_0$. 
Our model generates the terrain configurations for different parameter combinations shown in Fig. \ref{pics:terr_prop}.
% 
% The circle-puls notation in this figure is ambiguous. During the modulation of the latent, it works like the mathematical addition between a vector and a real number. But when applied to the raw vectors in training, I suspect, the circle-puls notation is something like concatenation. Therefore, I suggest using a black bar to replace the plus notation between o_t \tilde{z}_t and \hat{h}^f_t.
\begin{figure*}[h]
   \centering
    \includegraphics[width=0.80\textwidth, height=0.42\textwidth]{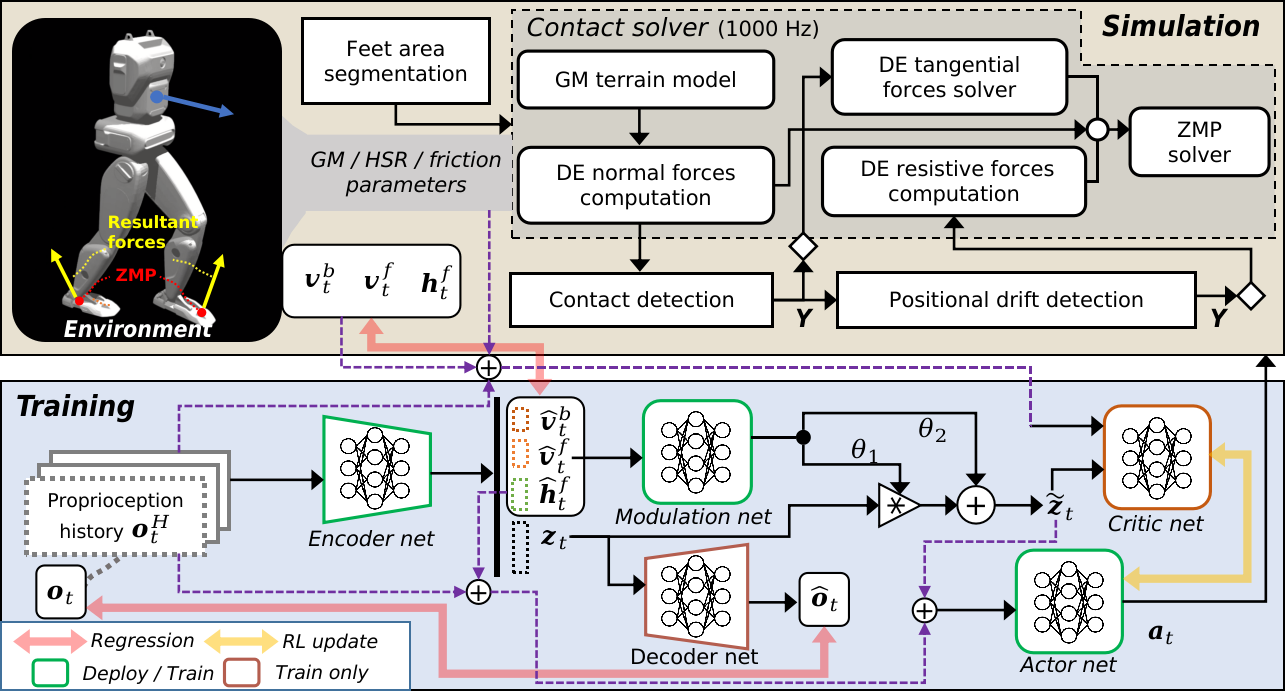}
   \caption{ Schematic of the simulation and training pipeline. 
   % The foot shape is segmented into discrete elements (DE) for spatially resolved contact force computation. 
   The contact solver integrates depth-dependent granular dynamics and drifting resistance to model heterogeneous foot-terrain interactions. In the asymmetric actor-critic (AAC) training pipeline, a VAE-based estimator reconstructs privileged states from proprioception, while the modulation net dynamically adjusts latent terrain representations using estimated feet velocity to enable implicit adaptation. }
   \label{pics:framework}
   \vspace{-1.5em}
\end{figure*}
% ***********************************************************************
% ****************************** SECTION 3 ******************************
% ***********************************************************************
\subsection{Whole-body Dynamics and Kinematic-aware Contact Solver} \label{GM_solver}
\subsubsection{System modeling} The vertical force computed via a discrete GM model determines the total contact count $N_c$, which initializes the tangential force solver. 
For all  $N_c$ contacts, the contact solver aims to drive the contact velocity to zero ($\boldsymbol{v}_{t+1}^{\{i\}} \rightarrow 0$) by iteratively adjusting tangential forces based on the predicted velocity deviation.
To establish this, we bridge the system dynamics with contact interactions through the floating-based formulation.

The bipedal robot, as depicted in Fig. \ref{pics:cover}, is essentially a poly-articulated floating-base system that can be modeled as a series of interconnected rigid bodies.
The resulting dynamics are therefore governed by the following Newton-Euler equation,
\begin{equation}
\boldsymbol{M}(\boldsymbol{q}) \dot{\boldsymbol{\nu}} + \boldsymbol{b}(\boldsymbol{q}, \boldsymbol{\nu}) = \boldsymbol{\tau} + \boldsymbol{J}_{c}^{T}(\boldsymbol{q}) \boldsymbol{F}_{c},
\end{equation}
where $\boldsymbol{M}$ is the mass matrix, $\boldsymbol{\tau}$ is a generalized force, and $\boldsymbol{q}, \boldsymbol{\nu} \in \mathbb{R}^{6+n_j}$ is the generalized coordinate and velocity, which involves the base linear velocity $\boldsymbol{v}_{b}$, angular velocity $\boldsymbol{\omega}_{b}$ and the joint velocity. 
$\boldsymbol{b}$ represents the nonlinear effects (i.e. Coriolis, centrifugal and gravitational terms).
By discretizing and mapping it to the world frame via jacobian $\boldsymbol{J}_{c}$, we obtain the velocity change at each contact point:
\begin{equation}  \label{eq:map}
\boldsymbol{J}_{c}\Delta \boldsymbol{\nu} = \Delta \boldsymbol{v} = \boldsymbol{J}_{c}\boldsymbol{M}^{-1}\boldsymbol{J}_{c}^{T} \boldsymbol{I}_{t} + \boldsymbol{J}_{c}\boldsymbol{M}^{-1}\Delta t(\boldsymbol{\tau}-\boldsymbol{b}),
\end{equation}
where $\boldsymbol{I}_{t}$ is the impulse generated by contact during interval $\Delta t$.
Next, to distinguish the rotational effects on different end-effectors $e$, we introduce the notation $\{e,i\}$ to represent elements. 
% Frictional effects are modeled as force-only interactions on each element, with negligible moments.
We then generalize Eq. \eqref{eq:map} to account for multiple contact interactions by superposing the velocity changes resulting from all active contacts. The contribution of $\boldsymbol{I}_{t}$ to both linear and angular velocity can be expressed as:
\vspace{-0.5em}
\begin{equation}\label{dyanmics_up}
\begin{split} 
    \boldsymbol{v}_{t+1}^{\{e, i\}} & = \boldsymbol{v}_{t}^{\{e,i\}} + \sum_{j=1}^{N_c} \boldsymbol{J}_{c,\mathcal{L}}^{\{e,i\}}  \boldsymbol{M}^{-1} \boldsymbol{J}_{c,\mathcal{L}}^{T\{e,j\}}\boldsymbol{I}_{t}^{\{e,j\}} \\
    & + \boldsymbol{J}_{c,\mathcal{L}}^{\{e,i\}}\boldsymbol{M}^{-1} \Delta t (\boldsymbol{\tau}- \boldsymbol{b})
\end{split}
\end{equation}
\vspace{-0.7em}
\begin{equation} \label{dyanmics_rot}
\begin{split} 
    \boldsymbol{\omega}_{t+1}^{\{e\}} & = \boldsymbol{\omega}_{t}^{\{e\}} + \sum_{j=1}^{N_c} \boldsymbol{J}_{c,\mathcal{R}}^{\{e\}}  \boldsymbol{M}^{-1} \boldsymbol{J}_{c,\mathcal{L}}^{T\{e,j\}}\boldsymbol{I}_{t}^{\{e, j\}} \\
    & + \boldsymbol{J}_{c,\mathcal{R}}^{\{e\}}\boldsymbol{M}^{-1} \Delta t (\boldsymbol{\tau}- \boldsymbol{b})
\end{split}
\end{equation}
where $\boldsymbol{J}_{c,\mathcal{L}}$ and $\boldsymbol{J}_{c,\mathcal{R}}$ are the translational and rotational Jacobians that respectively map generalized velocities to translational and angular motion at contact points. 
$\boldsymbol{\omega}_{t+1}^{\{e\}}$ is the shared angular velocity of the end-effector's body (COM-referenced).
It is noteworthy that in the Isaac Gym, only the Jacobian at the start of a rigid body, i.e., ankle point $\boldsymbol{J}_{A}$ is accessible. 
The Jacobian at each discrete element's center is obtained via the deduced rigid body transformation ($[\cdot]_{\times}$ denotes the skew matrix of the given vector).
\begin{align}
    \boldsymbol{J}^{\{i\}} = 
    \begin{bmatrix}
    \boldsymbol{I}_{3\times3} & - [\boldsymbol{p}^{\{i\}}]_{\times}\\
    0_{3\times3} & \boldsymbol{I}_{3\times3}
    \end{bmatrix}
    \boldsymbol{J}_{A}
\end{align}

\subsubsection{Physical constraint}
The predicted velocities of individual elements at time $t+1$ are calculated in isolation. 
% This independent processing inherently introduces discontinuities into the overall system dynamics. The issue becomes particularly pronounced when accounting for the physical interactions among the elements. 
% To guarantee the motion consistency in impulse updates, a kinematic constraint is formulated.
\textcolor{black}{This ignores physical interactions between elements, causing discontinuous motion. To ensure consistency, a kinematic constraint is applied during impulse updates.}
\textcolor{black}{Considering any two points $i$ and $j$ on the feet with a relative distance $\boldsymbol{r}^{\{ij\}}$,} the relative velocity error $\boldsymbol{\delta}_{rb}^{\{ij\}}$ between them can be expressed by the following formula:
\begin{equation} \label{eq:delta_rb}
    \boldsymbol{\delta}_{rb}^{\{ij\}} = \| \boldsymbol{v}_{t+1}^{\{i\}} - \boldsymbol{v}_{t+1}^{\{j\}} - \boldsymbol\omega_{t+1}  \times \boldsymbol{r}^{\{ij\}} \|,  \{\forall i, j\in N_c\}
\end{equation}
The average constraint violation $\boldsymbol{\delta}_{rb}^{\{i\}}$ is computed for element $i$ by considering its relative motion with respect to all preceding elements, quantifying the kinematic inconsistency at the current optimization step.
\begin{equation} \label{eq:delta_rb_i}
    \boldsymbol{\delta}_{rb}^{\{i\}} = \frac{1}{i-1}\sum_{j<i}\boldsymbol{\delta}_{rb}^{\{ij\}}
\end{equation}
% Moreover, the friction cone condition checks whether the tangential impulse of the current step $\boldsymbol{I}_{xy,t}^{\{i\}}$, plus the corrective terms required to drive the desired velocity to zero ($\beta \boldsymbol{v}_{t+1}^{\{i\}}$) and maintain kinematic coherency ($\gamma \boldsymbol{\delta}_{rb}^{\{i\}}$), lies within the friction cone formed by the normal impulse, i.e. $\mathcal{C} :  \mu \boldsymbol{I}_{z,t}^{\{i\}} > \| \boldsymbol{I}_{xy,t}^{\{i\}} - \gamma \boldsymbol{\delta}_{rb}^{\{i\}} -  \beta \boldsymbol{v}_{t+1}^{\{i\}} \|$. 
\textcolor{black}{Moreover, the friction cone condition checks whether the tangential impulse $\boldsymbol{I}_{xy,t}^{\{i\}}$ of the current step — plus two corrective terms — lies within the friction cone formed by the normal impulse. The corrective terms are: (1) $\beta \boldsymbol{v}_{t+1}^{\{i\}}$, to drive the desired velocity to zero; and (2) $\gamma \boldsymbol{\delta}_{rb}^{\{i\}}$, to maintain kinematic coherency. This constraint is written as $\mathcal{C} :  \mu \boldsymbol{I}_{z,t}^{\{i\}} > \| \boldsymbol{I}_{xy,t}^{\{i\}} - \gamma \boldsymbol{\delta}_{rb}^{\{i\}} -  \beta \boldsymbol{v}_{t+1}^{\{i\}} \|$, which can be finally formalized as}
\begin{equation} \label{eq:fric_combined}
\begin{cases}
    \boldsymbol{I}_{xy,t,next}^{\{i\}} = \boldsymbol{I}_{xy,t}^{\{i\}} - \gamma \boldsymbol{\delta}_{rb}^{\{i\}} - \beta \boldsymbol{v}_{t+1}^{\{i\}}, & \text{if } \mathcal{C} \\
    \boldsymbol{I}_{xy,t,next}^{\{i\}} = \mu \boldsymbol{I}_{xy,t}^{\{i\}} \frac{\boldsymbol{I}_{xy,t}^{\{i\}}- \gamma \boldsymbol{\delta}_{rb}^{\{i\}} - \beta \boldsymbol{v}_{t+1}^{\{i\}}}{\| \boldsymbol{I}_{xy,t}^{\{i\}} - \gamma \boldsymbol{\delta}_{rb}^{\{i\}}- \beta \boldsymbol{v}_{t+1}^{\{i\}} \|}, & \text{Otherwise}.
\end{cases}
\end{equation}
Together, these terms ensure that the solver balances local contact physics with consistency across the footplate, avoiding unphysical artifacts like partial slipping or motion tearing. 
The complete implementation of the contact solver is outlined in Algorithm 1.

% ****************************************************************
% *******************************************************************
% *******************************************************************

\begin{algorithm} 
\footnotesize
\caption{\footnotesize GM Terrain Model Multi-contact Solver at Timestep $t$}
\begin{algorithmic}[1]
\State \textcolor{blue}{\texttt{Input}}: \par  
$\boldsymbol{J}^{\{i\}}$, $\boldsymbol{M}$, $h_t^{\{i\}}$, $\dot{h}_t^{\{i\}}$, $\boldsymbol{s}_{t-1}^{\{i\}}$, $\boldsymbol{s}_t^{\{i\}}$, $\boldsymbol{v}_{t}^{\{i\}}$, $d_{D, t-1}^{\{i\}}$, $i=1,...,N$

\State \textcolor{blue}{\texttt{Initialization}}: \par 
\State $N_{c}=0$
\For{$i = 1 : N$}  % for 循环
    % \State $\tau_r^{\{i\}}  = \min(\tau_r^{\{i\}} + c_r [\boldsymbol{v}_{t}^{\{i\}}, \boldsymbol{v}_{t-1}^{\{i\}} < 0], 1.0)$
    \State $\boldsymbol{I}_{xy,t}^{\{i\}} = \{0, 0\}$, $\boldsymbol{I}_z^{\{i\}} = \Delta t \boldsymbol{F}_{GM}(h_t^{\{i\}}, \dot{h}_t^{\{i\}})$
    % \If {$\boldsymbol{I}_z^{\{i\}} > \boldsymbol{I}_{thres}$}  \hfill \textcolor{blue}{// \texttt{Contact detection}}
    %     \State $N_{c} = N_{c} + 1$
    % \Else 
    %     \State $d_{HSR, t-1}^{\{i\}} = 0$
    % \EndIf
    \State \texttt{contact-detection} ($N_c$, $\boldsymbol{I}_z^{\{i\}}$)
\EndFor

\State \textcolor{blue}{\texttt{Main}}: \par 

\While{$e \geq e_c$} \hfill \textcolor{blue}{// \texttt{Iterative solution loop}}
    \State $e = 0$
    \For{$i = 1 : N_c$}                
        \State $\boldsymbol{v}_{t+1}^{\{i\}}$ $\boldsymbol{\omega}_{t+1}$ = \texttt{dynamics-update}($N_c$, $\boldsymbol{v}_{t}^{\{i\}}$, $\boldsymbol{J}^{\{i\}}$, $\boldsymbol{M}$, $\boldsymbol{I}^{\{1:N_c\}}$ ) \hfill \textcolor{blue}{// \texttt{Eqs. (\ref{dyanmics_up}-\ref{dyanmics_rot})}}
        
        \State $\boldsymbol{\delta}_{rb}^{\{i\}}$ = \texttt{continuity-error}($\boldsymbol{v}_{t+1}^{\{i\}}$, $\boldsymbol{\omega}_{t+1}$) \hfill \textcolor{blue}{// \texttt{Eqs. (\ref{eq:delta_rb}-\ref{eq:delta_rb_i})}}
        
        \State $ \boldsymbol{I}_{xy,t,next}^{\{i\}}$ = \texttt{friction-constraint}($\boldsymbol{\delta}_{rb}^{\{i\}}$, $\boldsymbol{v}_{t+1}^{\{i\}}$, $\boldsymbol{I}^{\{i\}}$) \hfill \textcolor{blue}{// \texttt{Eq. (\ref{eq:fric_combined})}}
        
        \State $e = e + \| \boldsymbol{I}_{xy,t,next}^{\{i\}} - \boldsymbol{I}_{xy,t}^{\{i\}} \|$
        
        \State $\boldsymbol{I}^{\{i\}} = \{ \boldsymbol{I}_{xy,t,next}^{\{i\}}, \boldsymbol{I}_{z,t}^{\{i\}} \}$
        
    \EndFor
\EndWhile

\For{$i = 1 : N_c$}   \hfill \textcolor{blue}{// \texttt{Resistive force}}
    \State Calculate $d_{D, t}^{\{i\}} = d_{D, t-1}^{\{i\}} + \| \boldsymbol{s}_t^{\{i\}} - \boldsymbol{s}_{t-1}^{\{i\}} \|$
    \State Calculate $\boldsymbol{F}_{D}^{\{i\}}(d_{D}^{\{i\}}, h_t^{\{i\}}, \dot{\boldsymbol{v}}_{t}^{\{i\}})$ \hfill \textcolor{blue}{// \texttt{Eq. (\ref{eq:hsr_force})}}
\EndFor

\State \textcolor{blue}{\texttt{Return}}: \par 
$\boldsymbol{F}_{t}^{\{i\}} = \{ \boldsymbol{F}_{D,t}^{\{i\}} +  \frac{\boldsymbol{I}_{t,next}^{\{i\}}}{\Delta t}, \frac{\boldsymbol{I}_{z}^{\{i\}}}{\Delta t}  \}$
\end{algorithmic}
\end{algorithm}
% \vspace{-1.0em}

\subsubsection{Drifting Resistance}
\textcolor{black}{Additionally, a drift-resistive force counteracts horizontal motion, scaling with both penetration depth and horizontal displacement.}
We find that defining the travel distance as the accumulated path length from the starting intrusion position to the current position, rather than the direct Euclidean distance between these two points as in \cite{choi2023learning}, yields better performance, i.e. $d_{D} = \sum_{t} \| \boldsymbol{s}_t^{\{i\}} - \boldsymbol{s}_{t-1}^{\{i\}} \| 
$.
% \begin{equation} \label{eq:hsr_dist}
% d_{D} = \sum_{t} \| \boldsymbol{s}_t^{\{i\}} - \boldsymbol{s}_{t-1}^{\{i\}} \| 
% \end{equation}
% A buffer is set to record the intruded element's first intrusion till the current step. 
The drift-resistive force can be obtained in the following form
\begin{equation} \label{eq:hsr_force}
\boldsymbol{F}_{D} = -(k_{D} d_D h_t + b_D  \| \boldsymbol{v}_{t} \| ) \hat{\boldsymbol{v}}_{t}  
\end{equation}
where $\hat{\boldsymbol{v}}_{t}$ is the unit vector of $\boldsymbol{v}_{t}$, while $k_{D}$ and $b_{D}$ denote the stiffness and damping factor. Finally, we solve the zero moment points (ZMP) given the resultant force $\boldsymbol{F}_t = \sum_{i=1}^{N} \boldsymbol{F}_{t}^{\{i\}}$ and the calculated resultant torque $\boldsymbol{\tau}_t = \sum_{i=1}^{N} \boldsymbol{F}_{t}^{\{i\}} \times \boldsymbol{p}^{\{i\}}$. 
\vspace{-2pt}
\begin{equation}
    \boldsymbol{r}_{zmp} = [\boldsymbol{F}_{t}]_{\times}^{+} \cdot \boldsymbol{\tau}_t,
\end{equation}
\vspace{-2pt}
where $\boldsymbol{r}_{zmp}$ is the distance from the ZMP to the ankle points, and $[\boldsymbol{F}_{t}]_{\times}^{+}$ is the Moore-Penrose pseudoinverse of the skew matrix of the $\boldsymbol{F}_{t}$.

% ***********************************************************************
% ****************************** SECTION 4 ******************************
% ***********************************************************************
\subsection{Terrain-aware Reinforcement Learning Pipeline} \label{sec:training}
% ***********************************************************************
% ****************************** SECTION 1 ******************************
% ***********************************************************************
Our RL pipeline is depicted in Fig. \ref{pics:framework}. This control problem can be formulated as a Markov Decision Process (MDP) defined by $\{ \mathcal{S}, \mathcal{A}, r, p, \gamma\}$, where $\mathcal{S}$ is the state space and $\mathcal{A}$ is the action space. At state $\boldsymbol{s}_t$ of time step $t$, a policy $\pi$ performs an action $\boldsymbol{a}_t$ to forward the environment to the next state $\boldsymbol{s}_{t+1}$ with transition probability $p(\boldsymbol{s}_{t+1} | \boldsymbol{s}_{t}, \boldsymbol{a}_t)$ meanwhile receiving reward $r_t = r(\boldsymbol{s}_t, \boldsymbol{a}_t)$. 
RL aims to find the optimal parameter $\theta$ that maximizes the discounted return: $J(\theta) = \mathbb{E}_{\pi_{\theta}}\left[\sum_{t=0}^{\infty}(\gamma^{t}r_{t})\right]$.
% Consistent with this formulation, we provide detailed descriptions of each component.

% ****************************** SUBSECTION ******************************
\subsubsection{Observation-action space} \label{low_level_state} 
An asymmetric actor-critic framework \cite{pinto2017asymmetric} is employed where the actor and the critic receive different inputs (Fig. \ref{pics:framework}).
The actor takes $H$-step historical proprioceptive states $\boldsymbol{o}^{H}_{t}$ as its input, which includes base angular velocity $\boldsymbol{\omega}$, commanded velocity $\boldsymbol{v}^{cmd}$, projected gravity $\boldsymbol{g}$, joint angles $\boldsymbol{q}$, joint velocities $\boldsymbol{\dot{q}}$, and the action of the last step $\boldsymbol{a}_{t-1}$. 
In addition to this, the actor also receives a series of estimates of the robot's key states and a latent representation (Section \ref{seq:network}).
The critic additionally takes in the ground-truth states that are only accessible in simulation, as well as the real parameters of the terrain. 
The action $\boldsymbol{a}_t \in\mathbb{R}^{12}$ is chosen as the desired deviation of the joint angle from a time-invariant nominal pose $\boldsymbol{\mathring{q}}$, i.e. $\boldsymbol{a}_{t} = \boldsymbol{q}^{des}_{t} - \boldsymbol{\mathring{q}}$. 
% The joint-level PD controller generates a torque to track the final desired joint angle $\boldsymbol{q}^{des}_{t}$.
% i.e. $\boldsymbol{\tau} = \boldsymbol{K}_p \cdot (\boldsymbol{q}^{*}_{t} - \boldsymbol{q}_t) + \boldsymbol{K}_d \cdot ( - \boldsymbol{\dot{q}}_{t})$. 
% To mitigate the reality gap in real-world deployment, this motor actuation module is replaced by an actuator network trained with data collected from real machines, following the steps in our previous work \cite{luo2024moral}.

\begin{table}[h]
\caption{\textcolor{black}{Major reward terms for training policy. $(\cdot)^{cmd}$ and $(\cdot)^{des}$ is the desired and commanded physical values. $\mathcal{C}$ represents the feet contact states. $\mathbb{I}$ is the indicator function that returns 1 when the condition is true and 0 otherwise.} }
\vspace{-1.2em}
\label{reward_function}
\setlength\tabcolsep{3pt}  
\begin{center}
\begin{tabular}{c c c}
\hline
\textbf{Term} & \textbf{Equation}  & \textcolor{black}{\textbf{Weight}} \\
\hline
\\ [-1.5ex]
Lin. velocity tracking & $\exp\{-5 (\boldsymbol{v}_{xy}^{cmd}-\boldsymbol{v}_{xy})^2\}$  & 1.4\\ [+0.3ex]
Ang. velocity tracking &  $\exp\{-5(\boldsymbol{\omega}^{cmd}_{z}-\boldsymbol{\omega}_{z})^2\}$    & 1.1\\ [+0.3ex]
Feet clearance  &  $\|  \boldsymbol{h}^{f, des}_{t} - \boldsymbol{h}^{f}_{t}\} \| \cdot (1 - \mathcal{G}_{p})$   & 2.4\\ [+0.3ex]
Gait alignment & $ 1.3 \sum_{i=1}^{2} \mathbb{I}(\mathcal{C}_{i} = \mathcal{G}_{p,i}) - 0.3$ & 1.4\\ [+0.3ex]
Feet slippery speed & $\boldsymbol{v}^{f}_{t} \cdot \mathcal{C}$ & -0.1 \\ [+0.3ex]
% Feet distance & \makecell{$\exp \{-100 \cdot (d - d_{min}) \}$ \\ $+ \exp \{ -100 \cdot (d - d_{max})\}$}  & 0.2 \\ [+0.3ex]
% Default joint pose & \makecell{$\exp\{ -100( \| \boldsymbol{q}_{1,2} - \overset{\circ}{\boldsymbol{q}}_{1,2} \| ) \}$ \\ $+ (\boldsymbol{q} - \overset{\circ}{\boldsymbol{q}})^2$} & 0.8 \\ [+0.3ex]
% Base height & $h_{b} - \sum (\boldsymbol{h}_{f}\cdot \mathcal{C}) / \mathcal{C} $ & 0.2 \\ [+0.3ex]
Base acceleration & $\exp\{ -3 \| \boldsymbol{v}^{b}_{t} - \boldsymbol{v}^{f}_{t-1} \| \}$ & 0.2 \\ [+0.3ex]
Action smoothness & \makecell{$(\boldsymbol{a}_{t} - \boldsymbol{a}_{t-1})^{2}$ \\ $+ (\boldsymbol{a}_{t} - 2\boldsymbol{a}_{t-1} + \boldsymbol{a}_{t-2})^{2}$} & -0.003 \\ [+0.3ex]
% Joint torque & $\boldsymbol{\tau}^2$ & $-1 \times 10^{-10}$ \\ [+0.3ex]
% Joint acceleration & $(\dot{\boldsymbol{q}}_{t} - \dot{\boldsymbol{q}}_{t-1})^{2}$ & $-5 \times 10^{-9}$ \\ [+0.3ex]
% Collision & $n_{collision}$ & -1.0 \\ [+0.3ex]

\hline

\end{tabular}
\end{center}
% \vspace{-3.5em}
\end{table}

\begin{figure}[b]
   \centering
\includegraphics[width=0.49\textwidth, trim=1 10 1 1,clip]{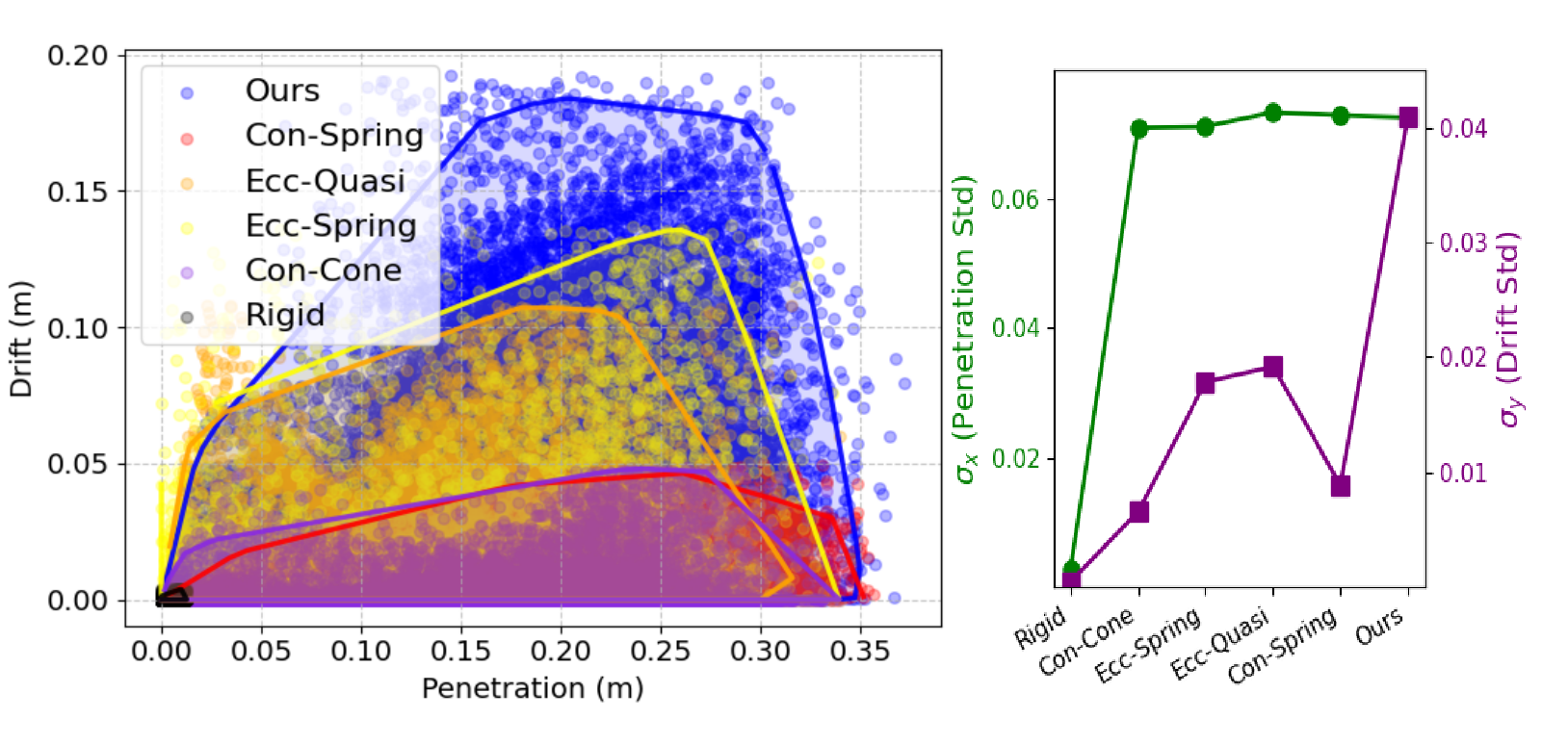}
   \caption{\textcolor{black}{The distribution of positional drift versus penetration for different methods and their standard deviations along the x and y directions.} }
   \label{pics:penetration}
   % \vspace{-1.2em}
\end{figure}

% ****************************** SUBSECTION ******************************
\subsubsection{Reward engineering} \label{seq:rew}
% This controller aims to learn a periodic gait defined by a sinusoidal function to track the commanded velocity. The gait $\mathcal{G}_{p}$ defines a schedule of contact events that vary over time. 
% We encourage the current contacts to align closely with the predefined gait schedule. 
Our controller learns a velocity-tracking periodic gait $\mathcal{G}_{p}$ defined by a sinusoidal function that schedules time-varying contact events. We optimize foot contacts to closely follow this gait schedule.
Some regularization terms are introduced to make the motion style more aesthetically pleasing and ensures stability during deployment. 
% These include penalizing the inappropriate distance between the feet and the deviation from the default joint pose $\overset{\circ}{\boldsymbol{q}}$.
% Other regularization ensures stability during deployment. For example, large fluctuations in actions and base height, as well as excessive torque are penalized.
The detailed reward terms and their weights can be found in Table \ref{reward_function}.

% ****************************** SUBSECTION ******************************
\subsubsection{Terrain Estimator and Modulation Net} \label{seq:network}
Previous work demonstrates the benefits of using latent variables for reasoning about the terrain \cite{lee2020learning} and an online estimation of robot states to improve robustness \cite{ji2022concurrent}. 
To this end, we introduce a multi-head variational autoencoder (VAE) \cite{higgins2017beta} as the first adaptive module. It performs estimation of base velocity, feet velocities and feet heights $\{ \boldsymbol{v}^{b}_{t}, \boldsymbol{v}^{f}_{t}, \boldsymbol{h}^{f}_{t}\}$, and outputs a latent variable $\boldsymbol{z}_t$ encoding implicit robot-environment information. The network is jointly optimized by regression and PPO policy loss:
\vspace{-3.0pt}
\begin{equation}
\begin{split} 
    Loss & = \epsilon \cdot loss_{reg} + (1-\epsilon) \cdot loss_{policy} \\
    loss_{reg} & = \mathcal{L}_{VAE} + \sum_{x \in \{\boldsymbol{v}^{b}_{t}, \boldsymbol{v}^{f}_{t}, \boldsymbol{h}^{f}_{t}\}} \mathbf{MSE}(x, \hat{x}) 
\end{split}
\end{equation}
\vspace{-2.0pt}
where hyperparameter $\epsilon \in [0,1]$ ($\epsilon = 0.3$ in our case). 
The loss $\mathcal{L}_{VAE}$ aims to reconstruct the observation to 
$\hat{\boldsymbol{o}}_t$ and minimize the KullbackLeibler (KL) divergence. 
$p(\boldsymbol{z}_t | \boldsymbol{o}^{H}_{t})$ is the posterior distribution of $\boldsymbol{z}_t$ given $\boldsymbol{o}^{H}_{t}$. 
$p(\boldsymbol{z}_t)$ is the prior distribution of $\boldsymbol{z}_t$ which is assume to be $\mathcal{N}(0, \mathbf{I})$.
\vspace{-3.0pt}
\begin{equation}
    \mathcal{L}_{VAE} = \mathbf{MSE}(\boldsymbol{o}_t, \hat{\boldsymbol{o}}_t) + \varphi \mathbf{D}_{KL} (p(\boldsymbol{z}_t | \boldsymbol{o}^{H}_{t}) \| p(\boldsymbol{z}_t))
\end{equation}
% \vspace{-5.0pt}
The reduced stiffness of compliant terrain alters contact restitution, causing foot velocity patterns distinct from high-stiffness surfaces.
These dynamics are encoded into the latent variable  $\boldsymbol{z}_t$ to enhance terrain awareness.
We propose to use a modulation network \cite{perez2018film} to form another adaptive module. This net dynamically adjusts $\boldsymbol{z}_t$ through a parametric transformation governed by $\boldsymbol{v}^{f}_t$.  This transformation layer processes $\boldsymbol{v}^{f}_t$ to generate two adaptive scaling factors $(\theta_{1}, \theta_{2})$, which then perform an affine operation on $\boldsymbol{z}_t$ expressed as $\tilde{\boldsymbol{z}}_{t} = \theta_1 \cdot \boldsymbol{z}_t + \theta_2$.
% \begin{equation}
    
% \end{equation}
% \vspace{-1.0em}
\begin{figure}[b]
   \centering
\includegraphics[width=0.49\textwidth, height=0.27\textwidth, trim=10 10 1 1,clip]{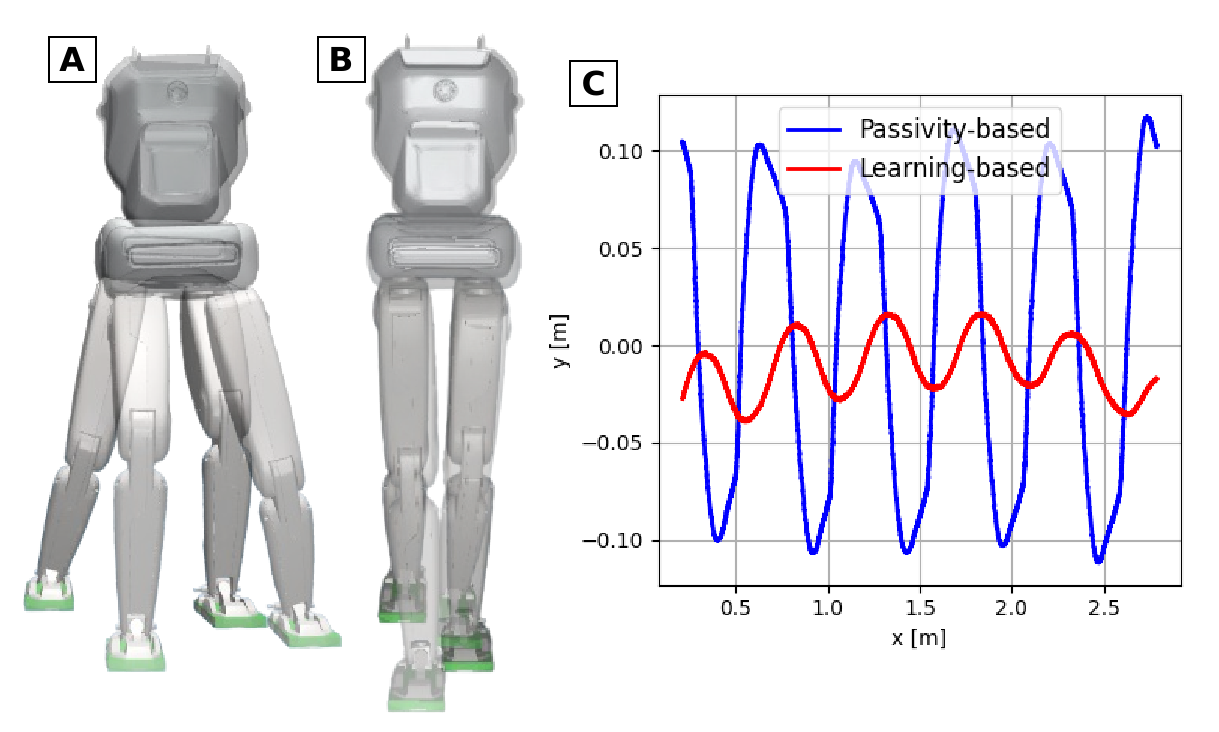}
   \caption{\textcolor{black}{Performance of  (\textbf{A}) the passivity-based controller and (\textbf{B}) the learning-based controller. (\textbf{C}) Center of Mass (CoM) trajectory comparison.}}
   \label{pics:com_manu}
\end{figure}

% ***************************************************************************
% ***************************************************************************
% ****************************** S E C T I O N ******************************
% ***************************************************************************
% ***************************************************************************
\section{Experimental Results}\label{Experiments}
% ***********************************************************************
% ****************************** SECTION 1 ******************************
% ***********************************************************************
\subsection{Implementation details}
% ****************************** SUBSECTION ******************************
% \subsubsection{Simulation}
We train the controller on 4096 agents in parallel on the Isaac Gym simulator for 20000 episodes. 
% An episode is terminated and reset under specific termination criteria when the robot body is in contact with the ground. 
% Once terminated, the robot is reset to the state sampled from the reference dataset $\mathcal{M}$. 
The linear velocity command in the forward direction is set as $[0.0, 1.2]\ m/s$
% ranging from low to high speed, 
, and the angular velocity command as $[-1.5, 1.5]\ rad/s$. 
Different contact scenarios on deformable terrain are modeled through randomized parameters, which are listed alongside training parameters in Table \ref{rand_parameters}.
% The representative randomized parameters are listed in Table 
Our pipeline consists of five MLP components: an encoder, a decoder, a modulation net, an actor, and a critic, which are updated concurrently by the gradients of corresponding losses.
% All of them are designed as multi-layer perceptrons (MLPs) with Exponential Linear Units (ELU) as the activation function. As mentioned in Section 3, they are updated concurrently by the gradients of corresponding losses.
% 所有的net are updated concurrently，但是如section 3所提到的那样，由不同的loss的gradient更新。
The entire training is performed on a desktop PC with an NVIDIA RTX 4080 GPU, which costs approximately 42.4 hours.

\begin{figure}[t]
   \centering
\includegraphics[width=0.50\textwidth, trim=8 1 1 1,clip]{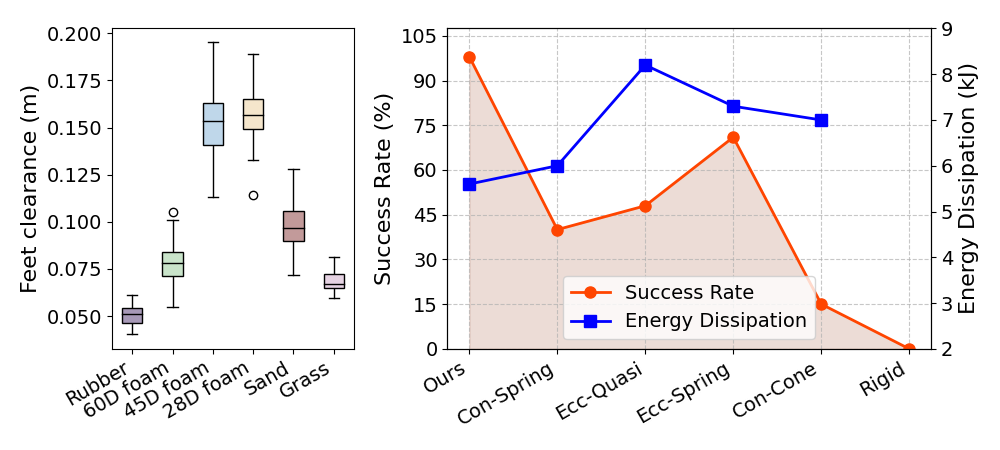}
   \caption{ \textcolor{black}{MILD controller's feet clearance across all terrain types (\textbf{Left}) and policy performance of the compared contact models on 25D foam (\textbf{Right}).} }
   \label{pics:energy_all}
   \vspace{-0.8em}
\end{figure}

\begin{table}[h]
% \vspace{-1.0em}
\caption{Randomization range of critical parameters}
\label{morph_table}
\begin{center}
\vspace{-1.0em}
\begin{tabular}{c c c c}
\hline
Category &  Parameters & Range & Unit\\
\hline
\\ [-2.1ex]

\multirow{8}*{Modeling} & Shear band angle $\alpha$   & [0.698, 1.310]   & $Rad$ \\
   % &  Ground Friction & [0.25, 1.75]   & - \\
& Compact rate $\nu$ & [0.8, 2.4]   & - \\
& Surr. mass scale $c_g$ & [-1.4, 4.6]   & - \\
& Inertial drag scale $c_d$ & [-6.0, 28.0]   &  - \\
& Volume fraction $\phi$ & [0.50, 0.64]   &  - \\
& Friction coefficient $\mu$ & [0.2, 1.3]  & - \\
& Pen. stiffness $k_{pen}$ & [8, 1400]  & $N/m$ \\
& Conical stress $\sigma_{cone}$ & [0.01, 0.24] & $10^{6}$ $N/m^{3}$ \\

\hline

\multirow{6}*{Training} & Added base mass   & [-4, 4]   & $Kg$ \\
   % &  Ground Friction & [0.25, 1.75]   & - \\
& Center of mass offset & [-6, 6]   &  $mm$ \\
& Motor latency & [1, 10]  & ms \\
& Motor offset & [-0.035, 0.035]  & $Rad$ \\
& IMU latency  & [1, 10] & $ms$ \\
& Command latency & [1, 10] & $ms$ \\

% \hline
% \\ [-2.1ex]
%    \multirow{5}*{High-level}  
%    & Gate center $x$ &  [1.2, 6.0] &  m \\
%    & Gate center $y$ &   [-1.8, 1.8] &  m \\
%    & Gate center $z$ &  [0.5, 0.63] & m \\
%    & Gate size $l^{out}_{G}$ &  [0.7, 1.0] &  m \\
%    & Gate thickness &  [0.02, 0.1] &  m \\
\hline

\end{tabular}
\end{center}
\label{rand_parameters}
% \vspace{-1.5em}
\end{table}

% \subsubsection{Hardware}
As shown in Fig. \ref{pics:cover}, the hardware validation test is carried out on the EngineAI SA01 bipedal robot \cite{engineAI}, which has a total of 12 degrees of freedom (DOF), with 6 DOF for each leg.
We employ a PD controller as the joint-level tracking of the desired joint angles. It operates at a frequency of 100 Hz, with $K_p=[50, 50, 70, 70, 20, 20], K_d=[5.0, 5.0, 7.0, 7.0, 0.2, 0.2] $ for the joints on each leg.
% We export the deployed policy from Isaac Gym and then reconstruct it via ONNXRuntime (Cross-platform inference accelerator). 

% ***********************************************************************
% ****************************** SECTION 2 ******************************
% ***********************************************************************
\subsection{Simulation Experiments} 
\label{sec:sim}
To assess the advantages of our approach, we compare it with existing state-of-the-art methods for deformable terrain modeling. The data is collected from the policies trained with the same episodes on their respective models:

\begin{itemize}[leftmargin=10pt]
    
    \item \textbf{Con-Spring} \cite{lynch2020soft}\cite{lynch2024efficient}: This model characterizes foot-substrate interaction as a concentric spring system that generates unidirectional penetration forces.
    \item \textbf{Con-Cone} \cite{choi2023learning}: \textcolor{black}{A developing cone model captures the added mass effect of the underlying substrate through concentric interaction at the geometric center.}
    \item \textbf{Rigid}: The baseline policy that is trained via the simulation's built-in rigid contact terrain.
    % \item \textbf{Ours}:  The proposed deformable surface eccentric interaction modeling.
    \item \textbf{Ecc-Quasi} (RFT) \cite{li2013terradynamics} \cite{xiong2017stability}: Uniformly distributed contact stress independent of penetration depth, neglecting inertial drag from substrate mass variation.
    \item \textbf{Ecc-Spring} \cite{lynch2020soft}\cite{lynch2024efficient}: The surface-type foot is discretized into multiple unidirectional spring elements.
\end{itemize}
% \vspace{-1.5em}
% \begin{figure}[t]
%    \centering
% \includegraphics[width=0.50\textwidth, trim=10 1 1 1,clip]{picture/baseline/energy.png}
%    \caption{Energy dissipation of our model and Ecc-Spring (Unit: kJ) }
%    \label{pics:energy}
%    \vspace{-1.5em}
% \end{figure}

% \vspace{-1.0em}
\begin{figure}[t]
   \centering
\includegraphics[width=0.53\textwidth, trim=1 1 1 10,clip]{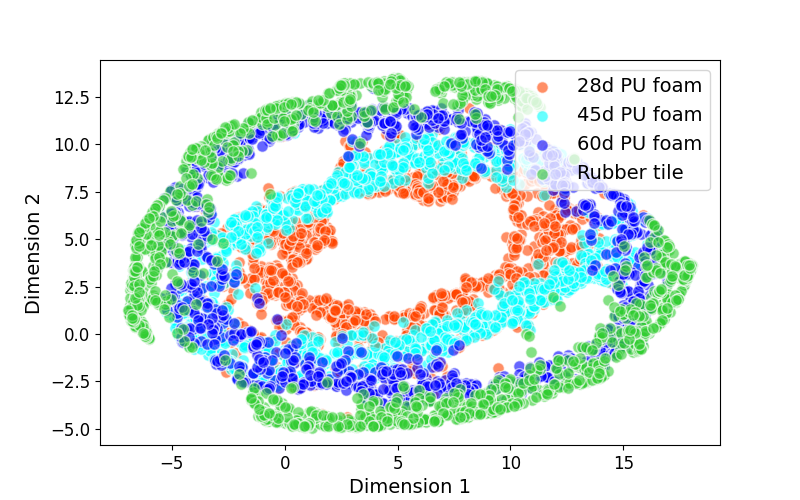}
   \caption{ The t-SNE classification of the encoded variable $\tilde{\boldsymbol{z}}_{t}$ across terrains with varying stiffness levels, where the dataset comprises 15-second walking trials on each terrain type.
 }
   \label{pics:tsne}
   \vspace{-0.8em}
\end{figure}

It is worthwhile to mention that \textbf{Ecc-Spring} is implemented to ensure comparison fairness, where \textbf{Con-Spring} is extended to the eccentric intrusion scenarios. As depicted in Fig. \ref{pics:penetration}, the two concentric single-point-contact cases exhibit the most limited performance (after rigid surface), as they fail to capture complex edge interactions and severely underestimate contact area.
In contrast, our approach achieves deeper penetration (+38 \% vs. Ecc-Spring) and longer slip distances, demonstrating broader contact interaction characteristics.
It outperforms the other two surface-expanded eccentric compression models because it fully accounts for substrate mass variation during penetration at different compression points, leading to a broad range of inertial drag. 
The improvements are more beneficial for the RL agent by helping the agent witness a wider range of randomization during training, enhancing the policy's adaptability in real-world situations.
% Our model Particularly superior in modeling edge-region drag for asymmetric contact scenarios.

\textcolor{black}{We further compare our method against a passivity-based whole-body controller \cite{mesesan2019dynamic} designed for compliant terrain, which requires computationally expensive online optimization to prioritize stability.
In contrast, MILD encodes robust dynamics through pre-optimized RL policies, resulting in significantly smaller Center of Mass (CoM) oscillations and greater stability during dynamic walking (Fig. \ref{pics:com_manu}).}
% This difference arises because the model-based control methods prioritize stability through computationally expensive online optimization, whereas learning-based approach encodes robust dynamics through pre-optimized policies, resulting in more natural and stable motion.

% We proceed to evaluate the energy dissipation of the two top-performing models from the robot's perspective. The metric is computed as the cumulative actuators' power expenditure  $\int_T | \boldsymbol{\tau} \cdot \dot{\boldsymbol{q}} | dt$ over a period of time.
% % As shown in Fig. \ref{pics:energy}, the robot exhibits substantially higher energy consumption at high velocities and on compliant substrates. Notably, our model demonstrates superior energy efficiency across all tested conditions.
% \textcolor{black}{As shown in Fig. \ref{pics:energy}, although energy use rises at high speeds or on soft terrain, our approach demonstrates superior efficiency across all tested conditions.}
% \textcolor{black}{This advantage arises from its high-fidelity representation of substrate dynamics—such as mass redistribution and local deformation—which reduces wasteful energy expenditure.}

% ***********************************************************************
% ****************************** SECTION 5 ******************************
% ***********************************************************************
\subsection{Real-World Experiments} \label{section: real-world}
\begin{figure*}[t]
   \centering
    \includegraphics[width=1.0\textwidth, height=0.57\textwidth, trim=1 1 1 1,clip]{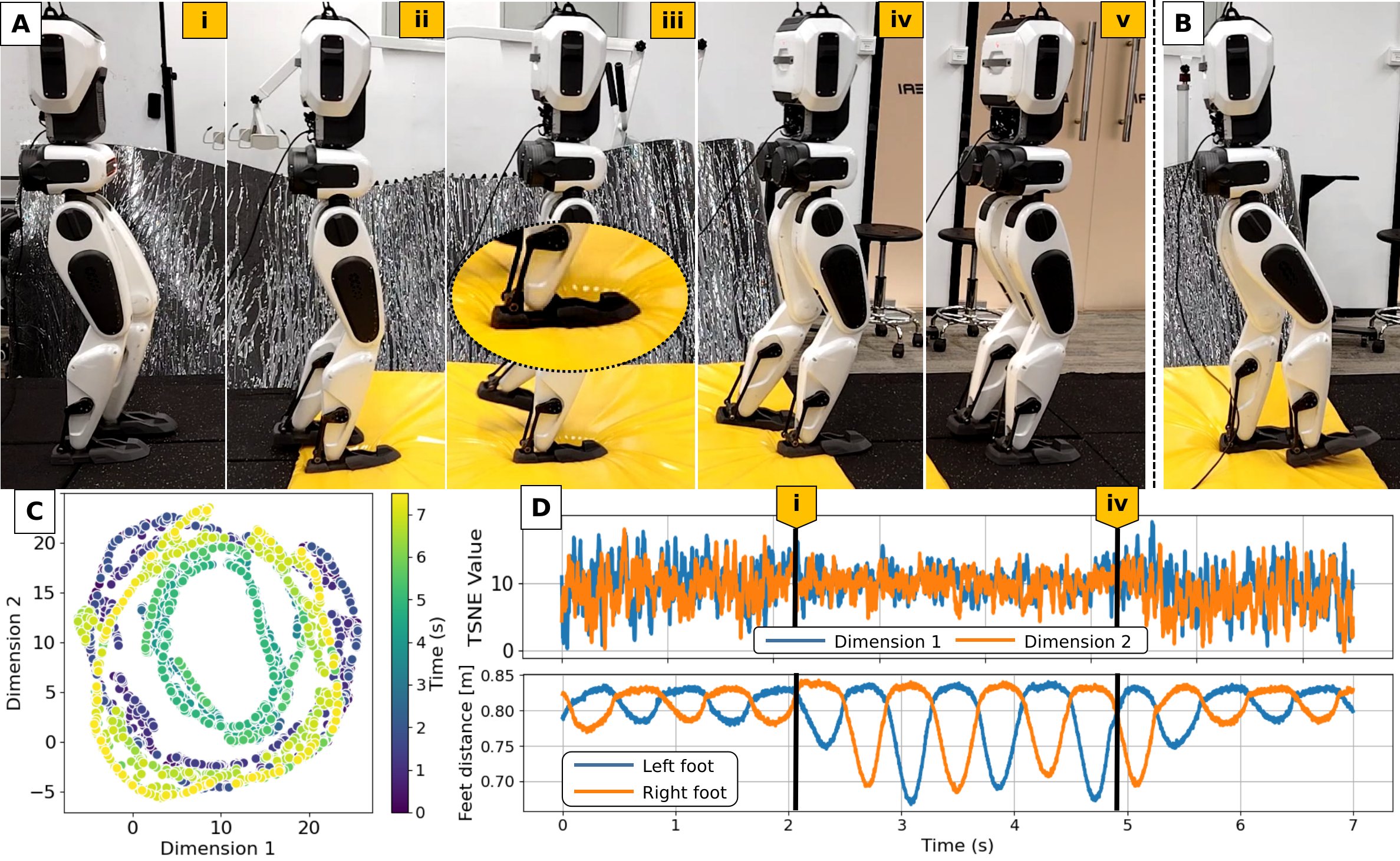}
   \caption{Robot motion analysis during sudden transition to foam penetration while moving forward (\textbf{A}) and backward (\textbf{B}), with corresponding t-SNE visualization of modulated latent space variations (\textbf{C}-\textbf{D}). The latent representations show distinct clustering patterns during terrain transitions, demonstrating the controller's real-time adaptation to varying surface compliance. }
   \label{pics:vis}
   \vspace{-1.2em}
\end{figure*}
% ***********************************************************************
% ***********************************************************************
\subsubsection{Implicit terrain adaptability}
Different from prior work testing on unquantified grass/sand, we evaluate system performance on standardized materials with defined stiffness grades. \textcolor{black}{Specifically, we consider six distinct terrains, including four manufactured surfaces—rubber tiles, 60d, 45d, and 28d polyurethane (PU) foam—and two natural surfaces: grass and sand, as shown in Fig. \ref{pics:cover}. }
The rubber tile provides slight deformability through its porous structure composed of granular elastomeric particles, while the PU foams exhibit a stiffness gradient based on their density grades, where smaller values indicate softer, more malleable surfaces. 
\textcolor{black}{Our controller demonstrates clear adaptability across these terrains (Fig. \ref{pics:energy_all}), automatically increasing foot clearance on softer surfaces to compensate for ground deformation.
Compared to other approaches, our fine-grained contact modeling achieves the highest success rate and energy efficiency.
}

To analyze the system's terrain implicit perception, we visualize the affine-transformed latent state $\tilde{\boldsymbol{z}}_{t}$ using t-SNE, as shown in Fig. \ref{pics:tsne}. 
The visualization reveals distinct clusters corresponding to the terrain's stiffness-deformability spectrum, with softer terrains occupying cluster interiors and stiffer terrains distributed peripherally. This clustering demonstrates our model's ability to encode terrain information that enables adaptive behavior generation across deformation levels.

% ***********************************************************************
% ***********************************************************************
\subsubsection{Comparative analysis}
This capability is verified through extensive trials across seven terrain types (including rigid planes). The robot successfully completes 10 forward/backward walking cycles at 1.2 m/s on each terrain without a single failure. 
For comparison, we also deploy controllers trained with baseline contact models (Section \ref{sec:sim}) on 45d foam. Due to the restricted training domains for contact conditions, these controllers adopt overly conservative gaits characterized by insufficient foot clearance during swing phases. 
This results in either joint limit violations or stability loss - failures that highlight the importance of proper leg lifting. 
% 我们展示了机器人在四种不同地面上的performance, including rubber tile, 60d PU foam, 45d PU foam, 28d PU foam。Different from the previous work, test on某些特定的软地形，我们systematically 通过海绵的指标来探究从硬到软地形的适应性（介绍这些deformable 地形的特征）Rubber tile颗粒橡胶，有一些表面的软度。值得指出的是foam前面的数字代表了海绵的密度从而代表了其对应的软硬程度。

\textcolor{black}{We further compare against two state-of-the-art learning-based methods: \textit{HT-2} \cite{radosavovic2024learning}, a transformer-based history-conditioned policy, and \textit{Clock} \cite{singh2024robust}, a clock-signal guided controller. 
Both methods achieve generalization through terrain geometry randomization on rigid surface without explicit modeling of terrain mechanics.}
\textcolor{black}{As shown in Table \ref{tab:velocity_tracking}, MILD shows an overwhelming performance compared to these methods in energy efficiency (COT), velocity tracking, and joint torque across various commanded velocities.
This fact suggests that the integration of a high-fidelity compliant contact model significantly enhances the controller's performance.}

\begin{table}[H]
\centering
\color{black} % 将整个表格设置为红色
\caption{\textcolor{black}{Performance comparison. Cost of Transport is calculated as $\text{COT} = \text{Power}/ (\text{Weight}\cdot \text{Velocity})$ with $\text{Power} = \sum_{actuators} \| \tau \dot{q} \|$ }}
\label{tab:velocity_tracking}
\renewcommand{\arraystretch}{1.2} % 调大行间距
\begin{tabular}{|c|c|c|c|c|c|} % 左右加竖线，列格式改为全竖线分割
\hline
\multicolumn{2}{|c|}{\makecell{Commanded Velocity $\boldsymbol{v}_x$(m/s)}} & 0.3 & 0.6 & 0.9 & 1.2 \\ % 表头左右竖线对齐
\hline
\multirow{3}{*}{\makecell{Cost of Transport \\(Hardware)}} 
& \textit{MILD}       & $\textbf{0.82} $ & $\textbf{0.75} $ & $\textbf{0.78} $ & $\textbf{0.83} $ \\
\cline{2-6} % 补充MILD与HT-2之间的横线
& \textit{HT-2}       & $1.05 $ & $0.96 $ & $1.20 $ & $1.08$ \\
\cline{2-6} % 补充HT-2与Clock之间的横线
& \textit{Clock}      & $1.38$ & $1.30$ & $1.35$ & $1.42$ \\
\hline
\multirow{3}{*}{\makecell{Maximum torque (Nm) \\(Hardware)}} 
& \textit{MILD}       & $\textbf{58.2}$ & $\textbf{63.2}$ & $\textbf{73.3}$ & $\textbf{88.4}$ \\
\cline{2-6} % 补充MILD与HT-2之间的横线
& \textit{HT-2}       & $62.9$ & $75.8$ & $96.0$ & $91.6$ \\
\cline{2-6} % 补充HT-2与Clock之间的横线
& \textit{Clock}      & $67.6$ & $82.5$ & $102.8$ & $115.2$ \\
\hline
\multirow{3}{*}{\makecell{Measured Velocity (m/s) \\(Simulation)}} 
& \textit{MILD}       & $\textbf{0.28}$ & $0.57$ & $\textbf{0.84}$ & $\textbf{1.15}$ \\
\cline{2-6} % 补充MILD与HT-2之间的横线
& \textit{HT-2}       & $\textbf{0.28}$ & $\textbf{0.61}$ & $0.98$ & $1.28$ \\
\cline{2-6} % 补充HT-2与Clock之间的横线
& \textit{Clock}      & $0.23$ & $0.54$ & $0.81$ & $1.09$ \\
\hline
\end{tabular}
\end{table}
\vspace{-0.5em}
% ***********************************************************************
% ***********************************************************************
\subsubsection{Online identification and transition}
The system demonstrates emergent terrain adaptation when the robot transitions between surfaces of varying stiffness without prior knowledge of this change (Fig. \ref{pics:vis}).
As observed in the linked video, when the robot moves from a high-stiffness to a low-stiffness surface using an initially conservative gait, the controller automatically increases stride length to maintain dynamic stability.
This adaptation results from variations in the latent state, which in turn modulate the generated motion.
The t-SNE visualization clearly captures this process, exhibiting distinct latent space trajectories during the complete rubber $\Rightarrow$ foam $\Rightarrow$ rubber transition cycle.
Remarkably, our pipeline maintains effective adaptation even when encountering abrupt terrain changes that are never experienced during training, indicating that the latent representation learns fundamental physical properties of ground stiffness rather than merely memorizing specific training conditions.

% ***************************************************************************
% ***************************************************************************
% ****************************** S E C T I O N ******************************
% ***************************************************************************
% ***************************************************************************
\section{CONCLUSIONS}\label{conclutions and future works}

We presented a high-fidelity compliant contact model for bipedal locomotion on deformable terrain, focusing on footplates with substantial contact areas. The model captures key interaction dynamics—eccentric penetration, high-speed impacts, and spatially varying forces.
We paired this with an RL-based controller featuring online estimation and identification for compliance adaptation.
Through systematic comparisons against state-of-the-art models, our approach demonstrated superior performance in both simulated and real-world experiments. Hardware experiments validated the system's capability to rapidly adapt to stiffness changes, as demonstrated by seamless transitions between distinct deformable levels, while accurately classifying their mechanical properties. 

The proposed framework could be extended by incorporating visual terrain deformation perception for improved adaptability. Future work of this model may also explore multi-gait generalization, including jumping and running maneuvers, and their implementation on model-based control architectures.

\vspace{-0.5em}

%%%%%%%%%%%%%%%%%%%%%%%%%%%%%%%%%%%%%%%%%%%%%%%%%%%%%%%%%%%%%%%%%%%%%%%%%%%%%%%%
\bibliographystyle{IEEEtran} % 参考文献排版风格，这个是IEEE transaction的，其他可以自查
\bibliography{references.bib} % 导入bib，references为“references.bib"的文件名

@string{iros = "{IEEE/RSJ Int. Conf. on Robots and Intelligent Systems}"}

@string{icra = "{IEEE Int. Conf. on Robotics and Automation}"}

@book{poschel2005computational,
  title={Computational granular dynamics: models and algorithms},
  author={P{\"o}schel, Thorsten and Schwager, Thomas},
  year={2005},
  publisher={Springer Science \& Business Media}
}

@article{lynch2020soft,
  title={The soft-landing problem: Minimizing energy loss by a legged robot impacting yielding terrain},
  author={Lynch, Daniel J and Lynch, Kevin M and Umbanhowar, Paul B},
  journal={IEEE Robotics and Automation Letters},
  volume={5},
  number={2},
  pages={3658--3665},
  year={2020},
  publisher={IEEE}
}

@inproceedings{vasilopoulos2014compliant,
  title={Compliant terrain legged locomotion using a viscoplastic approach},
  author={Vasilopoulos, Vasileios and Paraskevas, Iosif S and Papadopoulos, Evangelos G},
  booktitle={2014 IEEE/RSJ International Conference on Intelligent Robots and Systems},
  pages={4849--4854},
  year={2014},
  organization={IEEE}
}

@article{li2013terradynamics,
  title={A terradynamics of legged locomotion on granular media},
  author={Li, Chen and Zhang, Tingnan and Goldman, Daniel I},
  journal={science},
  volume={339},
  number={6126},
  pages={1408--1412},
  year={2013},
  publisher={American Association for the Advancement of Science}
}

@article{ji2022concurrent,
  title={Concurrent training of a control policy and a state estimator for dynamic and robust legged locomotion},
  author={Ji, Gwanghyeon and Mun, Juhyeok and Kim, Hyeongjun and Hwangbo, Jemin},
  journal={IEEE Robotics and Automation Letters},
  volume={7},
  number={2},
  pages={4630--4637},
  year={2022},
  publisher={IEEE}
}

@article{lee2020learning,
  title={Learning quadrupedal locomotion over challenging terrain},
  author={Lee, Joonho and Hwangbo, Jemin and Wellhausen, Lorenz and Koltun, Vladlen and Hutter, Marco},
  journal={Science robotics},
  volume={5},
  number={47},
  pages={eabc5986},
  year={2020},
  publisher={American Association for the Advancement of Science}
}

@article{choi2023learning,
  title={Learning quadrupedal locomotion on deformable terrain},
  author={Choi, Suyoung and Ji, Gwanghyeon and Park, Jeongsoo and Kim, Hyeongjun and Mun, Juhyeok and Lee, Jeong Hyun and Hwangbo, Jemin},
  journal={Science Robotics},
  volume={8},
  number={74},
  pages={eade2256},
  year={2023},
  publisher={American Association for the Advancement of Science}
}

@article{makoviychuk2021isaac,
  title={Isaac gym: High performance gpu-based physics simulation for robot learning},
  author={Makoviychuk, Viktor and Wawrzyniak, Lukasz and Guo, Yunrong and Lu, Michelle and Storey, Kier and Macklin, Miles and Hoeller, David and Rudin, Nikita and Allshire, Arthur and Handa, Ankur and others},
  journal={arXiv:2108.10470},
  year={2021}
}

@article{pinto2017asymmetric,
  title={Asymmetric actor critic for image-based robot learning},
  author={Pinto, Lerrel and Andrychowicz, Marcin and Welinder, Peter and Zaremba, Wojciech and Abbeel, Pieter},
  journal={arXiv preprint arXiv:1710.06542},
  year={2017}
}

@article{chen2024identifying,
  title={Identifying Terrain Physical Parameters from Vision-Towards Physical-Parameter-Aware Locomotion and Navigation},
  author={Chen, Jiaqi and Frey, Jonas and Zhou, Ruyi and Miki, Takahiro and Martius, Georg and Hutter, Marco},
  journal={IEEE Robotics and Automation Letters},
  year={2024},
  publisher={IEEE}
}

@article{lynch2024efficient,
  title={Efficient, Responsive, and Robust Hopping on Deformable Terrain},
  author={Lynch, Daniel J and Pusey, Jason L and Gart, Sean W and Umbanhowar, Paul B and Lynch, Kevin M},
  journal={IEEE Transactions on Robotics},
  year={2024},
  publisher={IEEE}
}

@inproceedings{xiong2017stability,
  title={A stability region criterion for flat-footed bipedal walking on deformable granular terrain},
  author={Xiong, Xiaobin and Ames, Aaron D and Goldman, Daniel I},
  booktitle={2017 IEEE/RSJ International Conference on Intelligent Robots and Systems (IROS)},
  pages={4552--4559},
  year={2017},
  organization={IEEE}
}

@inproceedings{todorov2012mujoco,
  title={Mujoco: A physics engine for model-based control},
  author={Todorov, Emanuel and Erez, Tom and Tassa, Yuval},
  booktitle={2012 IEEE/RSJ international conference on intelligent robots and systems},
  pages={5026--5033},
  year={2012},
  organization={IEEE}
}

@MISC{coumans2021,
author =   {Erwin Coumans and Yunfei Bai},
title =    {PyBullet, a Python module for physics simulation for games, robotics and machine learning},
howpublished = {\url{http://pybullet.org}},
year = {2016--2021}
}

@article{katsuragi2007unified,
  title={Unified force law for granular impact cratering},
  author={Katsuragi, Hiroaki and Durian, Douglas J},
  journal={Nature physics},
  volume={3},
  number={6},
  pages={420--423},
  year={2007},
  publisher={Nature Publishing Group UK London}
}

@article{tsimring2005modeling,
  title={Modeling of impact cratering in granular media},
  author={Tsimring, LS and Volfson, D},
  journal={Powders and grains},
  volume={2},
  pages={1215--1223},
  year={2005},
  publisher={Balkema Rotterdam}
}

@article{aguilar2016robophysical,
  title={Robophysical study of jumping dynamics on granular media},
  author={Aguilar, Jeffrey and Goldman, Daniel I},
  journal={Nature Physics},
  volume={12},
  number={3},
  pages={278--283},
  year={2016},
  publisher={Nature Publishing Group UK London}
}

@inproceedings{hubicki2016tractable,
  title={Tractable terrain-aware motion planning on granular media: An impulsive jumping study},
  author={Hubicki, Christian M and Aguilar, Jeff J and Goldman, Daniel I and Ames, Aaron D},
  booktitle={2016 IEEE/RSJ International Conference on Intelligent Robots and Systems (IROS)},
  pages={3887--3892},
  year={2016},
  organization={IEEE}
}

@article{li2009sensitive,
  title={Sensitive dependence of the motion of a legged robot on granular media},
  author={Li, Chen and Umbanhowar, Paul B and Komsuoglu, Haldun and Koditschek, Daniel E and Goldman, Daniel I},
  journal={Proceedings of the National Academy of Sciences},
  volume={106},
  number={9},
  pages={3029--3034},
  year={2009},
  publisher={National Academy of Sciences}
}

@article{ding2013foot,
  title={Foot--terrain interaction mechanics for legged robots: Modeling and experimental validation},
  author={Ding, Liang and Gao, Haibo and Deng, Zongquan and Song, Jianhu and Liu, Yiqun and Liu, Guangjun and Iagnemma, Karl},
  journal={The International Journal of Robotics Research},
  volume={32},
  number={13},
  pages={1585--1606},
  year={2013},
  publisher={SAGE Publications Sage UK: London, England}
}

@article{treers2021granular,
  title={Granular resistive force theory implementation for three-dimensional trajectories},
  author={Treers, Laura K and Cao, Cyndia and Stuart, Hannah S},
  journal={IEEE Robotics and Automation Letters},
  volume={6},
  number={2},
  pages={1887--1894},
  year={2021},
  publisher={IEEE}
}

@inproceedings{di2018dynamic,
  title={Dynamic locomotion in the mit cheetah 3 through convex model-predictive control},
  author={Di Carlo, Jared and Wensing, Patrick M and Katz, Benjamin and Bledt, Gerardo and Kim, Sangbae},
  booktitle={2018 IEEE/RSJ international conference on intelligent robots and systems (IROS)},
  pages={1--9},
  year={2018},
  organization={IEEE}
}

@inproceedings{perez2018film,
  title={Film: Visual reasoning with a general conditioning layer},
  author={Perez, Ethan and Strub, Florian and De Vries, Harm and Dumoulin, Vincent and Courville, Aaron},
  booktitle={Proceedings of the AAAI conference on artificial intelligence},
  volume={32},
  number={1},
  year={2018}
}

@inproceedings{higgins2017beta,
  title={beta-vae: Learning basic visual concepts with a constrained variational framework},
  author={Higgins, Irina and Matthey, Loic and Pal, Arka and Burgess, Christopher and Glorot, Xavier and Botvinick, Matthew and Mohamed, Shakir and Lerchner, Alexander},
  booktitle={International conference on learning representations},
  year={2017}
}

@inproceedings{chen2024foot,
  title={Foot shape-dependent resistive force model for bipedal walkers on granular terrains},
  author={Chen, Xunjie and Anikode, Aditya and Yi, Jingang and Liu, Tao},
  booktitle={2024 IEEE International Conference on Robotics and Automation (ICRA)},
  pages={13093--13099},
  year={2024},
  organization={IEEE}
}

@online{engineAI,
  title        = {EngineAI},
  url          = {https://www.engineai.com.cn/product_three},
}

@article{henze2016passivity,
  title={Passivity-based whole-body balancing for torque-controlled humanoid robots in multi-contact scenarios},
  author={Henze, Bernd and Roa, Maximo A and Ott, Christian},
  journal={The International Journal of Robotics Research},
  volume={35},
  number={12},
  pages={1522--1543},
  year={2016},
  publisher={SAGE Publications Sage UK: London, England}
}

@inproceedings{mesesan2019dynamic,
  title={Dynamic walking on compliant and uneven terrain using DCM and passivity-based whole-body control},
  author={Mesesan, George and Englsberger, Johannes and Garofalo, Gianluca and Ott, Christian and Albu-Sch{\"a}ffer, Alin},
  booktitle={2019 IEEE-RAS 19th International Conference on Humanoid Robots (Humanoids)},
  pages={25--32},
  year={2019},
  organization={IEEE}
}

@inproceedings{singh2024robust,
  title={Robust Humanoid Walking on Compliant and Uneven Terrain with Deep Reinforcement Learning},
  author={Singh, Rohan P and Morisawa, Mitsuharu and Benallegue, Mehdi and Xie, Zhaoming and Kanehiro, Fumio},
  booktitle={2024 IEEE-RAS 23rd International Conference on Humanoid Robots (Humanoids)},
  pages={497--504},
  year={2024},
  organization={IEEE}
}

@article{radosavovic2024learning,
  title={Learning humanoid locomotion over challenging terrain},
  author={Radosavovic, Ilija and Kamat, Sarthak and Darrell, Trevor and Malik, Jitendra},
  journal={arXiv preprint arXiv:2410.03654},
  year={2024}
}

\end{document}